\documentclass{article}
\usepackage{spconf,amsmath,graphicx}

\usepackage{enumitem}
\setlist{nosep, leftmargin=14pt}

\usepackage{mwe} 

\usepackage{booktabs}
\usepackage{multirow}
\usepackage{url}

\title{Multi-Scale Context-Guided Lumbar Spine Disease Identification with Coarse-to-fine Localization and Classification}
\name{Zifan Chen$^1$ \qquad Jie Zhao$^2$ \qquad Hao Yu$^1$ \qquad Yue Zhang$^1$ \qquad Li Zhang$^{1,3}$\sthanks{Corresponding author: Li Zhang (zhangli\_pku@pku.edu.cn)}}

\address{$^1$ Center for Data Science, Peking University, China \\ $^2$ National Engineering Laboratory for Big Data Analysis and Applications, Peking University, China \\ $^3$ National Biomedical Imaging Center, Peking University, China}

\begin{document}
%
\maketitle
\begin{abstract}
Accurate and efficient lumbar spine disease identification is crucial for clinical diagnosis. However, existing deep learning models with millions of parameters often fail to learn with only hundreds or dozens of medical images. These models also ignore the contextual relationship between adjacent objects, such as between vertebras and intervertebral discs. This work introduces a multi-scale context-guided network with coarse-to-fine localization and classification, named CCF-Net, for lumbar spine disease identification. Specifically, in learning, we divide the localization objective into two parallel tasks, coarse and fine, which are more straightforward and effectively reduce the number of parameters and computational cost. The experimental results show that the coarse-to-fine design presents the potential to achieve high performance with fewer parameters and data requirements. Moreover, the multi-scale context-guided module can significantly improve the performance by 6.45\% and 5.51\% with ResNet18 and ResNet50, respectively. Our code is available at Github\footnote{https://github.com/czifan/CCFNet.pytorch}.
\end{abstract}

\section{Introduction}
\label{sec:introduction}

\begin{figure}[htb]
\begin{minipage}[b]{.40\linewidth}
  \centering
  \centerline{\includegraphics[width=3.2cm,height=1.5cm]{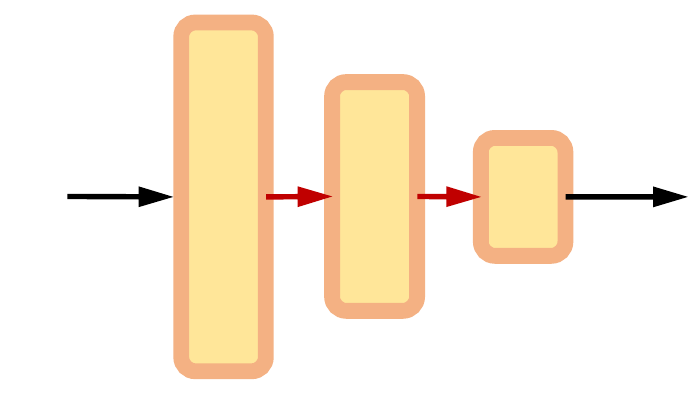}}
  \centerline{(a) CCF-Net}\medskip
\end{minipage}
\hfill
\begin{minipage}[b]{.60\linewidth}
  \centering
  \centerline{\includegraphics[width=3.4cm,height=1.5cm]{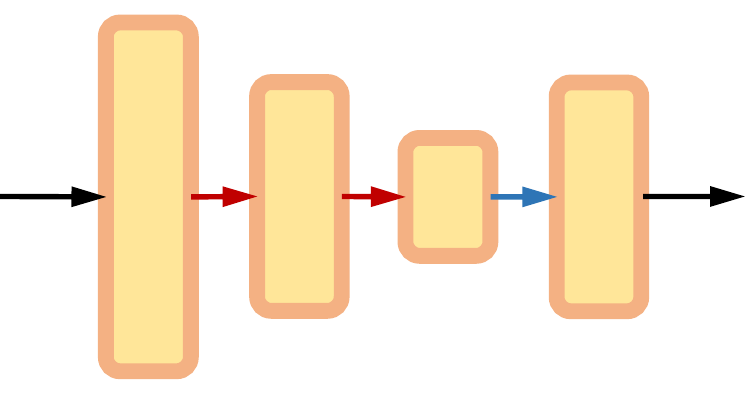}}
  \centerline{(b) Encoder-Decoder style}\medskip
\end{minipage}
\begin{minipage}[b]{1.0\linewidth}
  \centering
  \centerline{\includegraphics[width=7.8cm,height=3.0cm]{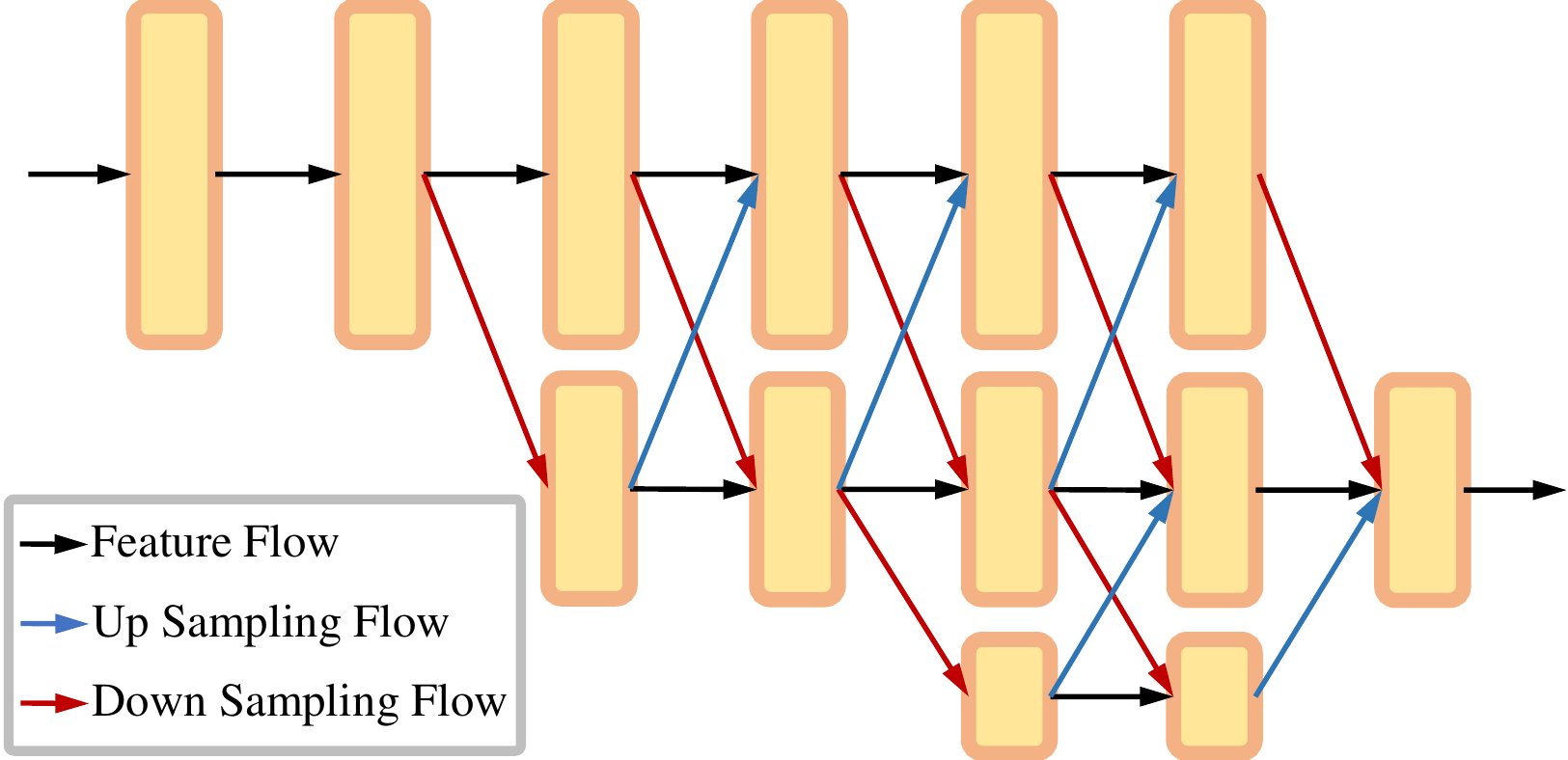}}
  \centerline{(c) High-Resolution style}\medskip
\end{minipage}
\caption{Illustration and comparisons of CCF-Net and other two classic style of frameworks for localizaiton.}
\label{fig:compare}
\end{figure}

\begin{figure*}[htb]
    \centering
    \includegraphics[width=0.9\textwidth]{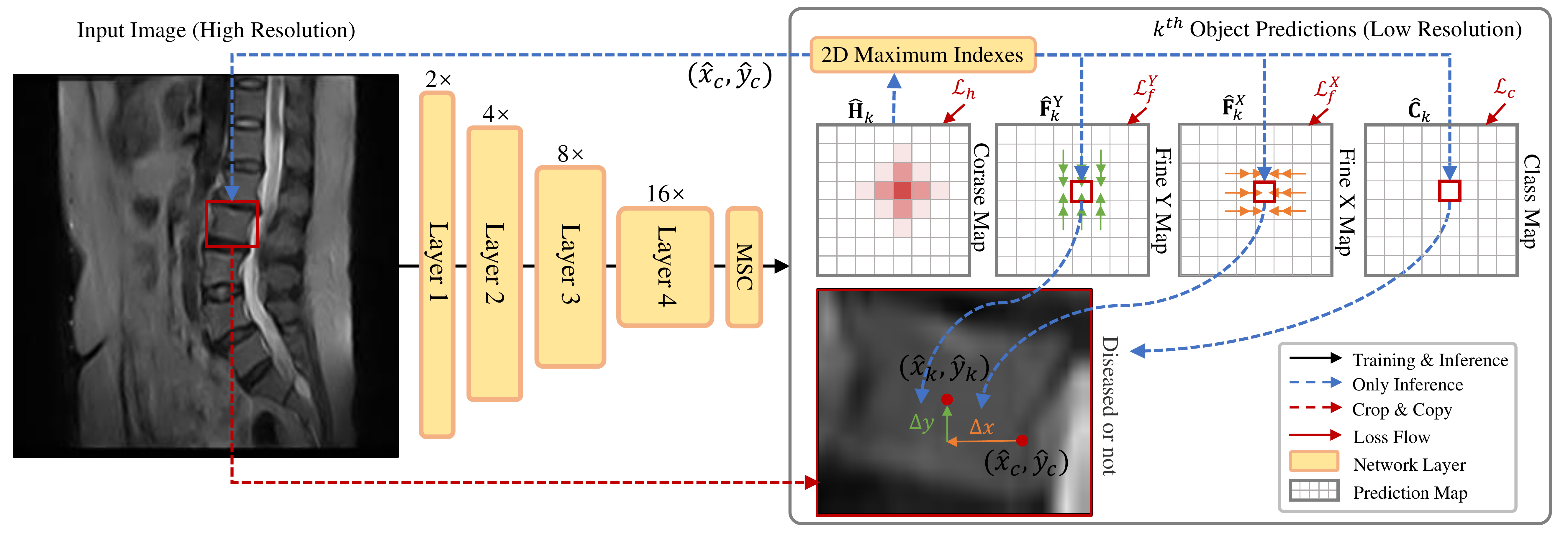}
    \caption{The framework of CCF-Net. The right sub-figure shows the parsing process for $k^{th}$ vertebra localization and classification, which is the same for other vertebras and intervertebral discs. (Best view in color)}
    \label{fig:framework}
\end{figure*}

Lumbar spine disease is one of the most prevalent health problems globally~\cite{wieser2011cost}. Magnetic resonance imaging (MRI) provides high contrast between soft tissue and bone, so it has been considered the gold standard to evaluate lumbar spine diseases~\cite{kim2020diagnostic}, such as pyramidal degeneration and lumbar intervertebral disc herniation. However, extensive work to locate and identify wood diseases from MRI creates a heavy workload for physicians. It is necessary to develop accurate and efficient computer-aided methods to reduce the workload of doctors.

Recently, various deep learning methods have been proposed to solve this problem~\cite{huang2020spine,dolz2018ivd,xiao2018simple,sun2019deep,sun2019high}. Typical localization methods include the Encoder-Decoder style (Fig.\ref{fig:compare} (b)) and the High-Resolution style (Fig.\ref{fig:compare} (c)). Some works~\cite{huang2020spine,xiao2018simple,newell2016stacked} use the Encoder-Decoder framework as their basic design. The encoder is used to extract deep features, while the decoder is responsible for fusing multi-scale features and predicting high-resolution heatmap for localization. On the other hand, the High-Resolution framework is first proposed by Sun et al.~\cite{sun2019deep} to maintain high-resolution features during feature extraction. Moreover, a series of improvements based on HRNet have been proposed, such as HRNetV2~\cite{sun2019high} and HigherHRNet~\cite{cheng2020higherhrnet}. However, both frameworks need complex architecture designs to predict high-resolution heatmap. Such designs would require more data to train an effective model with millions of parameters and much more computational cost to infer it during the actual radiological diagnosis. Besides, several works~\cite{dolz2018ivd,zhang2020sau} focus on segmenting intervertebral discs or vertebras to achieve localization, which needs many pixel-level annotations. However, collecting enough annotated data to train segmentation models in the medical field is challenging and even infeasible. 

It is worth mentioning that the relationship between adjacent vertebras and intervertebral discs is context-specific due to the sequential alignment for lumbar spine localization and disease classification. 
For an example, if the L4 vertebra is diseased, its adjacent vertebras (L3 and L5) and intervertebral discs (L3-L4 and L4-L5) are also highly likely to be diseased, according to the geometry and prior clinical knowledge~\cite{mayerhoefer2012quantitative}. Nevertheless, most of the existing methods~\cite{huang2020spine,li20183d,dolz2018ivd,xiao2018simple,sun2019deep,sun2019high,zhang2020sau} ignore this message in their models.

To solve the above two problems, we propose a multi-scale context-guided network with coarse-to-fine localization and classification, named CCF-Net. We divide the localization objective into two more straightforward parallel tasks (coarse and fine), which use a low-resolution heatmap (coarse map) to locate the approximate position of the targets and two corresponding offset maps (fine X/Y maps) to refine the precise coordinates of them. This design allows the overall model to use low-resolution predictions to reconstruct accurate coordinates at high-resolution without decoder or high-resolution path, as shown in Fig.\ref{fig:compare} (a). It has effectively reduced a large portion of parameters and computational costs. Besides, we introduce the multi-scale context-guided (MSC) module to model the contextual relationship between adjacent objects, and the experimental results show effectiveness. In summary, our main contributions are as follows: 
\begin{itemize}
    \item We design a coarse-to-fine localization framework with fewer parameters and data requirements.
    \item We introduce a multi-scale context-guided module in which the contextual relationship is embedded to significantly improve the performance of the detection tasks related to lumbar spine disease.
    \item We demonstrate the effectiveness of the proposed CCF-Net in a vast amount of experiments. 
\end{itemize}

\section{Method}
\label{sec:method}

\subsection{Framework}
As shown in Fig.~\ref{fig:framework}, the proposed CCF-Net takes a 2D slice image in MRI as its input and uses ResNet~\cite{he2016deep} as the encoder to extract features. We replace the last layer of ResNet as the multi-scale context-guided (MSC) module to model the contextual relationship between adjacent vertebras and intervertebral discs. For each object, CCF-Net would predict a low-resolution heatmap (coarse map) for approximate localization along with two offset maps (fine X/ Y map) for refining to obtain precise coordinates. Besides, a corresponding category map would be generated to predict the probability of disease at each location. We compute losses between these four maps and the ground truth maps converted by the original ground truth coordinate and disease category during training. Moreover, we would parse these maps during inference.

\subsection{Coarse-to-fine Localization}
\label{subsec:corse_to_fine}
The proposed CCF-Net directly uses the encoder to generate low-resolution coarse and fine maps with $H'\times W'$ without decoder and high-resolution path. Suppose the resolution of input image is $H\times W$, and the stride of the whole model is $S$ ($S=16$ in this work) with the relationship of $H'=\left \lfloor H/S \right \rfloor$ and $W'=\left \lfloor W/S \right \rfloor$. There are eleven objects (five vertebras and six intervertebral discs) $\left ((x_1,y_1,c_1),\cdots,(x_{11},y_{11},c_{11})\right )$ for localization and classification. For the $k^{th}$ object, the coarse map would be defined as
\begin{equation}
    \begin{aligned}
    \mathbf{H}_k(i,j)=exp\left ( \frac{-{\left (\left (t(i)-x_k\right )^2+\left (t(j)-y_k\right )^2\right )} }{2\sigma^2}\right ),
    \nonumber
    \label{eqn:corase_map}
    \end{aligned}
\end{equation} 
where $(i, j)$ is the position on the coarse map, and $\sigma$ is the parameter of a Gaussian blur. $t(\cdot)$ translates the low-resolution position to high-resolution space, which can be formatted as $t(z)=(z-A)\times S$, where $A$ is 0.5 as an aligning constant. At the same time, the fine maps can be expressed as
\begin{equation}
    \begin{aligned}
    \mathbf{F}^X_k(i,j)=\left ( t(i)-x_k\right ) / S,\quad
    \mathbf{F}^Y_k(i,j)=\left ( t(j)-y_k\right ) / S,
    \nonumber
    \label{eqn:fine_map}
    \end{aligned}
\end{equation} 
where we use a stride $S$ to scale the offset to make it easier to learn. For the category map, we set it to all ones if the current object is diseased; otherwise, to all zeros, as follows,
\begin{equation}
    \begin{aligned}
    \mathbf{C}_k(i,j)=\left\{\begin{array}{l} 1, if\ c_k\ is\ diseased \\ 0, otherwise\end{array}\right.
    \nonumber
    \label{eqn:category_map}
    \end{aligned}
\end{equation} 

\begin{figure}[htb]
    \centering
    \includegraphics[width=0.47\textwidth]{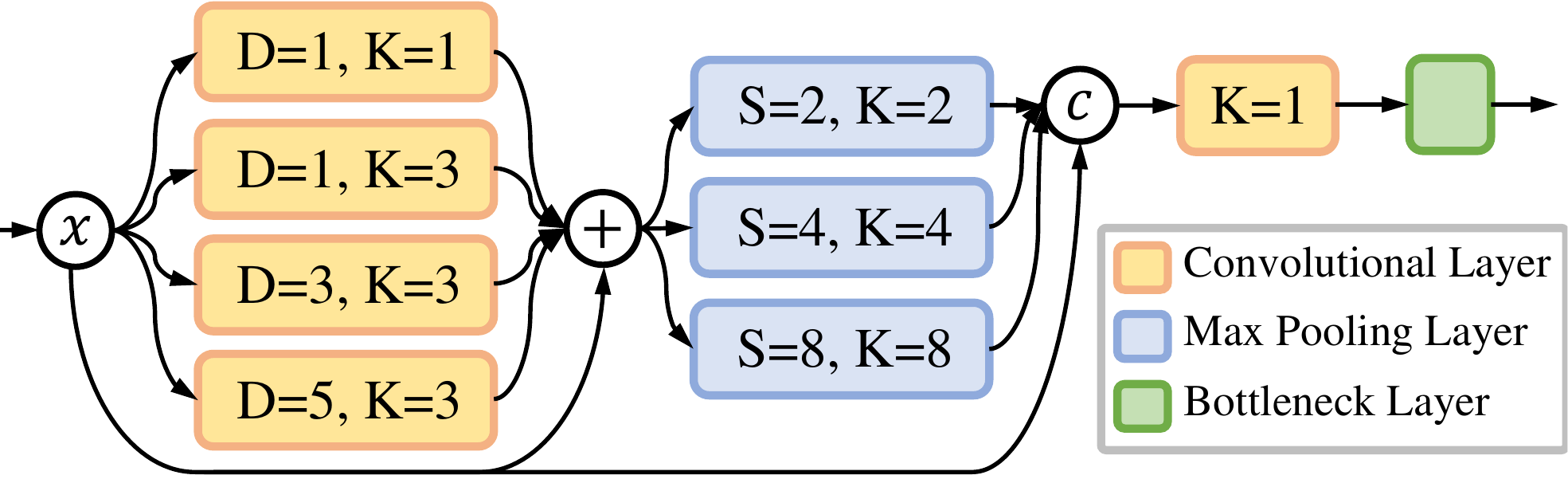}    \caption{Illustration of Multi-Scale Context-Guided (MSC) Module. ``D'', ``K'' and ``S'' indicates dilation size, kernel size, and stride, respectively. All strides that are not specified are set to 1 as default. Paddings are set to the corresponding value to keep the same resolution after convolution. ``$x$'' is the input feature map. ``$+$'' and ``$c$'' represent pixel-wise addition and channel-wise concatenation, respectively.}
    \label{fig:msc}
\end{figure}

\subsection{Multi-Scale Context-Guided Module}
The adjacent vertebras and intervertebral discs have a contextual relationship. Clinical evidence shows that if one of them is diseased, its adjacent objects are highly probable to be diseased~\cite{mayerhoefer2012quantitative}. We replace the last layer of ResNet~\cite{he2016deep} with the proposed Multi-Scale Context-Guided (MSC) module to model this relationship in feature space. As shown in Fig.~\ref{fig:msc}, the feature map will pass through a multi-dilation convolutional module and a multi-stride max-pooling in turn to capture feature relationships between different objects at different distances. And then, a $1\times 1$ convolutional layer and a Bottleneck introduced in ResNet~\cite{he2016deep} are used to fuse features and reduce channel dimension.

\subsection{Loss function}
During training, we compute the loss between four low-resolution prediction maps ($\hat{\mathbf{H}},\hat{\mathbf{F}}^X,\hat{\mathbf{F}}^Y,\hat{\mathbf{C}}$) and their corresponding ground truths ($\mathbf{H},\mathbf{F}^X,\mathbf{F}^Y,\mathbf{C}$). For the coarse map, we use mean square error as the objective, defined as
\begin{equation}
    \begin{aligned}
    \mathcal L_h=\frac{1}{N}\sum_k\sum_i\sum_j MSE(\hat{\mathbf{H}}_k(i,j),\mathbf{H}_k(i,j)),
    \nonumber
    \label{eqn:coarse_loss}
    \end{aligned}
\end{equation}
where $N$ is the total number of positions in $\hat{\mathbf{H}}$. For the fine map, we only compute the loss on the positions with high activation values in $\mathbf{H}$ and use smoothed L1 loss to suppress the influence of outliers during training~\cite{ren2015faster}. The objectives are
\begin{equation}
    \begin{aligned}
    \mathcal L^X_f=\frac{1}{M}\sum_{k}\sum_{(i,j)\in\Omega_k} Smooth_{L_1}(\hat{\mathbf{F}}^X_k(i,j),\mathbf{F}^X_k(i,j)), \\
    \mathcal L^Y_f=\frac{1}{M}\sum_{k}\sum_{(i,j)\in\Omega_k} Smooth_{L_1}(\hat{\mathbf{F}}^Y_k(i,j),\mathbf{F}^Y_k(i,j)),
    \nonumber
    \label{eqn:fine_loss}
    \end{aligned}
\end{equation}
where $\Omega_k=\{(i,j)|\mathbf{H}_k(i,j)\ge \tau \}$, and $\tau$ is set to 0.6 as the threshold. $M$ is the total number of positions in $\Omega$. For the category map, we compute the binary crossentropy loss for each position in $\Omega$, that is
\begin{equation}
    \begin{aligned}
    \mathcal L_c=\frac{1}{M}\sum_{k}\sum_{(i,j)\in\Omega_k} BCE(\hat{\mathbf{C}}_k(i,j),\mathbf{C}_k(i,j)).
    \nonumber
    \label{eqn:category_loss}
    \end{aligned}
\end{equation}
The overall objective can be summarized as
\begin{equation}
    \begin{aligned}
    \mathcal L=w_h\mathcal L_h+w_f(\mathcal L_f^X+\mathcal L_f^Y)+w_c\mathcal L_c,
    \nonumber
    \label{eqn:loss}
    \end{aligned}
\end{equation}
where $w_h,w_f,$ and $w_c$ are set to 1.0, 2.0, and 1e-3, respectively, to balance different parts of the objective.

\subsection{Inference}
For the inference of the model, we parse four prediction maps ($\hat{\mathbf{H}},\hat{\mathbf{F}}^X,\hat{\mathbf{F}}^Y,\hat{\mathbf{C}}$) to coordinates for localization and categories for classification. As shown in Fig.~\ref{fig:framework}, we take an example for $k^{th}$ object as follows. First, we find the position $(\hat{i},\hat{j})$ in $\hat{\mathbf{H}}_k$ with maximum activation value by the 2D maximum indexes module and then map $(\hat{i},\hat{j})$ to $(\hat{x}_c,\hat{y}_c)$ through ${t}(\cdot)$ as described as Sec.~\ref{subsec:corse_to_fine}. Second, we take $\hat{\mathbf{F}}^X_k(\hat{i},\hat{j})$ and $\hat{\mathbf{F}}^Y_k(\hat{i},\hat{j})$ as offsets on the X-axis and Y-axis, respectively, obtaining the precise coordinates of the $k^{th}$ object $(\hat{x}_k,\hat{y}_k)$. Third, we use $\hat{c}_k=\hat{\mathbf{C}}_k(\hat{i},\hat{j})$ as the probability of whether it is diseased.

\begin{table*}[htbp]
    \renewcommand\arraystretch{1.2}
    \centering
    \resizebox{0.95\textwidth}{!}{
    \begin{tabular}{l l c c c c c c c}
    \toprule
     \multirow{2}{*}{Method} & \multirow{2}{*}{Backbone} & \multirow{2}{*}{Params (M)} & \multirow{2}{*}{Flops (G)} & \multicolumn{2}{c}{Localization} & \multicolumn{2}{c}{Classification} & \multirow{2}{*}{Score} \\
      \cmidrule{5-6} \cmidrule{7-8}
    & & & & Disc & Vertebra & Disc & Vertebra & \\
    \midrule
    SimpleBaseline~\cite{xiao2018simple} & ResNet18 & 15.38 & 33.23 & 87.81\small{±2.88} & 87.68\small{±3.73} & 89.26\small{±1.47} & 71.71\small{±6.43} & 70.70\small{±5.51} \\
    SimpleBaseline~\cite{xiao2018simple} & ResNet18-MSC & 6.86 & 33.50 & 93.78\small{±2.89} & 93.64\small{±2.66} & 90.72\small{±2.10} & 76.97\small{±3.61} & 78.55\small{±2.87} \\
    SCN~\cite{huang2020spine} & ResNet18 & 26.57 & 42.73 & 88.56\small{±4.43} & 88.77\small{±4.20} & 89.26\small{±1.64} & 71.18\small{±4.13} & 71.18\small{±5.26} \\
    SCN~\cite{huang2020spine} & ResNet18-MSC & 11.62 & 45.43 & 92.79\small{±1.80} & 94.13\small{±1.59} & 90.16\small{±1.64} & 75.94\small{±5.07} & 77.64\small{±3.82} \\
    \midrule
    CCF-Net (ours) & ResNet18 & 9.51 & \textbf{11.20} & 89.69\small{±3.51} & 89.23\small{±4.42} & 89.32\small{±2.52} & 76.23\small{±2.99} & 74.05\small{±3.36} \\
                  & ResNet18-MSC & \textbf{4.83} & 12.00 & \textbf{94.75\small{±3.36}} & \textbf{94.71\small{±3.60}} & \textbf{90.88\small{±1.23}} & \textbf{79.16\small{±2.55}} & \textbf{80.50\small{±1.79}} \\
    \midrule
    \midrule
    SimpleBaseline~\cite{xiao2018simple} & ResNet50 & 34.00 & 51.64 & 89.33\small{±3.18} & 90.07\small{±3.52} & 90.21\small{±3.68} & 76.06\small{±2.71} & 74.59\small{±4.13} \\
    SimpleBaseline~\cite{xiao2018simple} & ResNet50-MSC & 45.33 & 124.80 & 94.53\small{±1.51} & 94.54\small{±1.34} & 90.56\small{±1.70} & 76.06\small{±2.76} & 78.77\small{±2.92} \\
    HRNet~\cite{sun2019deep} & W32 & 28.54 & 40.98 & \textbf{96.19\small{±1.26}} & 95.63\small{±1.53} & 89.68\small{±2.08} & \textbf{78.43\small{±2.70}} & 80.62\small{±2.22} \\
    \midrule
    CCF-Net (ours) & ResNet50 & 23.60 & \textbf{21.49} & 90.43\small{±3.46} & 90.24\small{±3.37} & 90.06\small{±0.72} & 76.35\small{±3.48} & 75.13\small{±1.80} \\
                  & ResNet50-MSC & \textbf{10.79} & 22.09 & 95.68\small{±3.98} & \textbf{96.41\small{±3.91}} & \textbf{90.59\small{±1.76}} & 77.46\small{±4.01} & \textbf{80.64\small{±1.65}} \\
    \bottomrule
    \end{tabular}
    }
    \caption{Comparisons of the results with four-fold cross validation. We report the recall score and the AUC score at the true points (within $6mm$ of the corresponding ground truth point) for localization and classification, respectively. The overall score is computed by the product of the average scores of localization and classification. The results for ``Localization'', ``Classification'' and ``Score'' are shown as the mean and the standard deviation of four-folds. Best results are shown in bold.}
    \label{tab:result}
\end{table*}

\section{Experiments}
\label{sec:experiments}

\subsection{Experimental Settings}
To evaluate the proposed CCF-Net's effectiveness, we run experiments on \emph{Spinal Disease Dataset} \cite{spinaldiseasedataset2020} on Tianchi Platform.
The dataset includes 201 patients, each with a T2 phase MRI sequence. One slice is labeled in each MRI sequence, including the localization and category for five vertebras (L1 to L5) and six intervertebral discs (L1-L2 to L5-S1). In our experiments, we use four-fold cross-validation and report the mean and standard deviation of the results to ensure a fair comparison. For all experiments, we uniformly adopt random rotation, random translation, random cropping, random flipping, random noising and resize the input image to $512\times 512$. We use PyTorch~\cite{paszke2019pytorch} to implement all methods and choose a learning rate of 3e-4 for the AdamW~\cite{loshchilov2018fixing} optimizer where the weight decay is set to 5e-4. The cosine annealing with warm restart method~\cite{loshchilov2016sgdr} is used as the learning rate decay scheduler. We train all models with 60 epochs and adopt a batch size of 32 for the ResNet18-based model and 12 for the rest.

\begin{figure}[htb]
    \centering
    \includegraphics[width=0.4\textwidth]{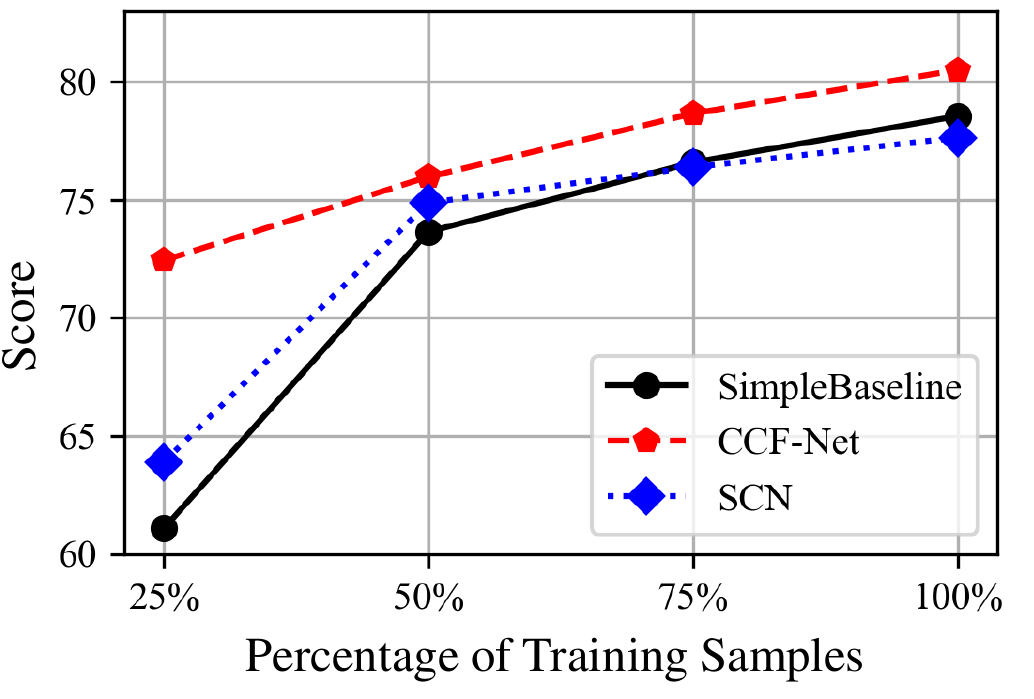}  
    \caption{The impact of the training sample size.}
    \label{fig:sample}
\end{figure}

\subsection{Comparison with Other Methods}
As shown in Table~\ref{tab:result}, we compare the proposed CCF-Net with the other three popular methods~\cite{huang2020spine,xiao2018simple,sun2019deep} with different backbones. CCF-Net can achieve potentiality performance with fewer parameters and computational cost. Specifically, SimpleBaseline~\cite{xiao2018simple} is the most similar architecture to ours. An essential difference between CCF-Net and SimpleBaseline is whether there is a decoder to reconstruct a high-resolution feature map. Compared with SimpleBaseline with the same backbone, the proposed coarse-to-fine design in CCF-Net can reduce about 38\% parameters and 66\% GFlops while bringing about 4.7\% performance improvement based on ResNet18. Moreover, replacing the last layer of ResNet with the MSC module can significantly improve the performance from 4.17\% to 6.46\% and reduce computational cost in most cases.

\subsection{The impact of the training sample size}
The more straightforward architecture with fewer parameters always means it is easier to learn and not easy to overfit. We conduct experiments to explore the performances of different methods with smaller training sets. As shown in Fig.~\ref{fig:sample}, the proposed CCF-Net can achieve potential performance with only about 10\% performance loss compared with SCN (18\%) and SimpleBaseline (22\%) on 25\% training data. As the training data increases, our method continuously improves performance and consistently outperforms the other two methods. The experimental result shows that the proposed CCF-Net is more suitable for medical imaging with fewer data.


\section{Conclusion}
\label{sec:conclusion}
Much of the existing localization methods predict high-resolution heatmaps with additional complex designs, such as encoder-decoder or high-resolution path, giving rise to more parameters and computational cost. In this work, we propose a multi-scale context-guided network with coarse-to-fine localization and classification, named CCF-Net, to improve the performance and reduce the complexity of network architecture. The advantages of the proposed method are confirmed in the experimental results on the challenging dataset. Moreover, the MSC module can effectively model the contextual relationship between vertebras and intervertebral discs. Furthermore, the straightforward architecture is more suitable for training on fewer medical images. This would be a fruitful area for further work to improve our method and extend it to other medical imaging localization tasks.

\section{Compliance with Ethical Standards}
This is a numerical simulation study for which no ethical approval was required.

\section{Acknowledgments}
\label{sec:acknowledgments}
This work is supported by the National Natural Science Foundation of China (12090022, 81801778, 11831002, 91959205), Beijing Natural Science Foundation (Z200015, Z180001), PKU-Baidu Foundation (2020BD027), and Pilot Project of Public Welfare Development Reform of Beijing-based Medical Research Institutes (2019-1). We also thank to Tianchi Platform, Wanli Cloud, and AllinMD Orthopaedics for providing the dataset.

\bibliographystyle{IEEEbib}
\bibliography{strings,refs}

\end{document}